\documentclass{article}

\usepackage{arxiv}

\usepackage[utf8]{inputenc} 
\usepackage[T1]{fontenc}    
\usepackage{hyperref}       
\usepackage{url}            
\usepackage{booktabs}       
\usepackage{amsfonts}       
\usepackage{nicefrac}       
\usepackage{microtype}      
\usepackage{lipsum}		
\usepackage{graphicx}
\usepackage{natbib}
\usepackage{doi}

\usepackage[utf8]{inputenc} 
\usepackage[T1]{fontenc}    
\usepackage{hyperref}       
\usepackage{url}            
\usepackage{booktabs}       
\usepackage{amsfonts}       
\usepackage{nicefrac}       
\usepackage{microtype}      
\usepackage{xcolor}
\usepackage{algorithm} 
\usepackage{amsmath}
\usepackage{algpseudocode} 
\usepackage{graphicx}
\usepackage{graphicx}
\usepackage{subcaption}

\title{Uncovering Drift in Textual Data: An Unsupervised Method for Detecting and Mitigating Drift in Machine Learning Models}

\author{ Saeed Khaki\\
	Amazon\\
	\texttt{sakhaki@amazon.com} \\
	\And
	Akhouri Abhinav Aditya\\
	Amazon \\
	\texttt{akhoura@amazon.com} \\
	\And
	Zohar Karnin\\
	Amazon \\
	\texttt{zkarnin@amazon.com} \\
	\And
	Lan Ma\\
	Amazon \\
	\texttt{mamlm@amazon.com} \\
	\And
	Olivia Pan\\
	Amazon \\
		\texttt{olipan@amazon.com} \\
	\And
	Samarth Marudheri Chandrashekar\\
	Amazon \\
	\texttt{smarudh@amazon.com} \\
}

\renewcommand{\shorttitle}{Uncovering Drift in Textual Data}

\hypersetup{
pdftitle={Uncovering Drift in Textual Data},
pdfsubject={q-bio.NC, q-bio.QM},
pdfauthor={saeed khaki},
pdfkeywords={Unsupervised drift detection, text data, machine learning},
}

\begin{document}
\maketitle

\begin{abstract}
 Drift in machine learning refers to the phenomenon where the statistical properties of data or context, in which the model operates, change over time leading to a decrease in its performance. Therefore, maintaining a constant  monitoring process for machine learning model performance is crucial in order to proactively prevent any potential performance regression. However, supervised drift detection methods require human annotation and consequently lead to a longer time to detect and mitigate the drift. In our proposed unsupervised drift detection method, we follow a two step process. Our first step involves encoding a sample of production data as the target distribution, and the model training data as the reference distribution. In the second step, we employ a kernel-based statistical test that utilizes the maximum mean discrepancy (MMD) distance metric to compare the reference and target distributions and estimate any potential drift. Our  method also identifies the subset of production data that is the root cause of the drift. The models retrained using these identified high drift samples show improved performance on online customer experience quality metrics.
\end{abstract}


\section{Introduction}

In the fast-paced world of big data, where the amount of information being generated is constantly growing, efficient data analytics and machine learning techniques are essential to help us make informed decisions. However, with the rapid emergence of new products, markets, and customer behaviors, a new challenge arises - the problem of drift in data. This occurs when the statistical properties of the data being used change over time in unforeseen ways. If left unchecked, data drift can render past data irrelevant, leading to poor decision outcomes. As a result, drift has become a significant obstacle for many data-driven and machine learning systems. In this dynamic and ever-evolving environment, finding ways to provide reliable and accurate data-driven predictions and decision-making capabilities is crucial \citep{lu2018learning,gemaque2020overview}.

Drift detection is the process of monitoring the performance of a machine learning model over time and detecting when the model's behavior deviates from its original distribution. There are two general methods for drift detection: supervised and unsupervised \citep{gemaque2020overview}. Supervised drift detection involves using labeled data to detect changes in the model's performance. This method assumes that there is a labeled dataset available that represents the original distribution of the data. The labeled data is used to train the model, and the performance of the model is measured on new labeled data as it becomes available. If the performance of the model deviates significantly from its original performance, the model is flagged for drift \citep{lu2018learning,maciel2015lightweight,gonccalves2014comparative,iwashita2018overview}. On the other hand, unsupervised drift detection involves monitoring the model's behavior without the use of labeled data. This method assumes that there is no labeled dataset available that represents the original distribution of the data. Instead, the model's behavior is monitored using statistical techniques such as change point detection or anomaly detection also distance between two distributions. These techniques compare the model's behavior over time and look for significant deviations that could indicate drift \citep{friedrich2023unsupervised, de2019learning,gemaque2020overview}. 

Despite supervised drift detection being accurate, it is not often used in practice compared to unsupervised drift detection methods because if labeled data is not available, the method cannot be used. Additionally, labeled data can be expensive and time-consuming to obtain, which can make supervised drift detection impractical in some situations \citep{gemaque2020overview}. In the literature, many studies have used these approaches to detect drift. \cite{harel2014concept} proposed a supervised method for identifying drift in data streams through the examination of empirical loss of learning algorithms. This approach involves extracting statistics from the distribution of loss by using resampling to repeatedly reuse the same data. \cite{maciel2015lightweight} proposed a new supervised ensemble classifier called drift detection ensemble (DDE), which is designed to improve the effectiveness of three different concept drift detectors. DDE combines the warnings and drift detections from these three detectors using various strategies and configurations to achieve better performance than the individual methods alone. \cite{li2019faad} proposed an unsupervised approach for detecting anomalies in multidimensional sequences in data streams that are susceptible to concept drift, by utilizing a feature selection algorithm based on mutual information and symmetric uncertainty information. The proposed approach, called FAAD, consists of three different algorithms and aims to analyze the ordering relationship in sequences to detect anomalies. \cite{costa2018drift} used an unsupervised drift detection approach, DDAL, that utilizes active learning to select the most significant instances for monitoring concept drift based on density variation. The proposed method involves two phases, where the first phase generates a classifier using reference data and the second phase involves drift detection, reaction, and classification for new batches of data based on the occurrence of concept drift. \cite{lughofer2016recognizing} introduces two techniques to handle concept drift in scenarios with few labeled instances or no labeled instances, one based on active learning filters and the other on analyzing classifier output certainty distribution. \cite{haque2016sand} proposed a semi-supervised framework for detecting drift and novel classes in data streams using an ensemble of kNN classifiers, with outlier detection and concept change detection modules, and incorporating data self-annotation.

In this paper, we present a novel unsupervised approach for detecting drift in unstructured text data used in machine learning models, without requiring human annotation. Our proposed method offers several key novelties which are as follows:

\begin{enumerate}

    \item Our proposed algorithm is a highly versatile, unsupervised drift detection method that can be applied to any machine learning model for both performance regression detection and mitigation in unstructured text data without the need for human input.
    
    \item Our method includes a mitigation strategy for addressing model performance regression. It offers a fast and reliable solution for improving model performance when high levels of drift are present in production data.
    
    \item We demonstrate the effectiveness of our approach in detecting and mitigating performance regression in a real-world application. By leveraging our novel method, we were able to achieve improved model performance, highlighting the practical value of our approach.
    
\end{enumerate}

\section{Methodology}

In this paper, we introduce an unsupervised drift detection method for unstructured data, specifically text data utilized as input for machine learning models. Our proposed method involves the encoding of unstructured text data into vector representation, followed by a comparison of the encoded data to identify potential drift. To achieve this, we use a maximum mean discrepancy (MMD) distance \citep{gretton2012kernel} in a kernel-based statistical test, leveraging bootstrap to provide statistics that characterize the drift, such as the median and mean. Additionally, our approach identifies the subset of data likely to be the root cause of the drift. Subsequently, this subset of data can be used in the retraining of models to minimize performance regression.

Specifically, we show how our proposed drift detection method can be used for detecting drift in the inference data for a machine learning model that is in production. Leveraging our proposed method, we can systematically compare the model development data, including training and validation data, with the production data to identify potential drift. Although we demonstrate the proposed method for text data in this paper, it can be used for other data modalities, such as image and voice data.

Our method compares the input data of a machine learning model trained on $(X_{tr},Y_{tr})$ with that of $X_{prod}$, a subset of production data. However, since both the training and production data can be large, we divide them into mini-batches and compare those mini-batches to identify potential drift. To mitigate sampling bias, we first shuffle both the training and production data before dividing them into mini-batches. Given that our proposed method is designed to detect drift in text data, we leverage an encoder such as BERT \citep{devlin2018bert} to obtain fixed-sized embeddings of the text data. Specifically, we compute the average embedding for a mini-batch as its overall vector representation. Subsequently, we use these embeddings as input in our proposed drift detection algorithm, which is presented in Algorithm \ref{al:al1}.

We used the maximum mean discrepancy (MMD) test \citep{gretton2012kernel} in our proposed method to compare the training (reference) and production (target) distributions. MMD is a distribution-free approach that is particularly well-suited for high-dimensional spaces. It measures the difference between two probability distributions by evaluating the distance between their means in a reproducing kernel Hilbert space (RKHS). MMD's strength lies in its ability to capture complex, nonlinear relationships between variables and its ability to handle high-dimensional data. Therefore, we selected MMD as it can provide an accurate representation of the dissimilarity between the reference and target distributions.

The proposed method utilizes bootstrap to estimate the drift distribution by comparing the training and production distributions. Initially, both distributions are combined under the null hypothesis of no drift. Subsequently, a bootstrapped distribution is obtained by randomly sampling with replacement. Next, the method calculates the drift between the first and second halves of the bootstrapped samples. If the null hypothesis holds, and there is no difference between the distributions, then the amount of drift (measured by MMD) in the bootstrapped distributions should be minimal. Otherwise, a significant drift is observed. Furthermore, the likelihood of the drift can be estimated using bootstrap.

To mitigate the performance regression caused by drift, we propose a process in which we examine the estimated drift between mini-batches from our proposed algorithm. This enables us to identify the samples that exhibit the greatest deviation from the production data, thereby isolating the most significant sources of drift. These samples can then be reintroduced into the training data, allowing the model to be refreshed and better equipped to handle future changes in the data distribution. This approach not only improves the model's overall performance, but also ensures its continued relevance and accuracy in dynamic, real-world settings.

\begin{algorithm}[h]
	\caption{Proposed Drift Detection Method} 
	\begin{algorithmic}[1]
\State \textbf{Result}: Estimated Drift, Samples Causing Drift
 \State \textbf{Inputs:}

     \State $D_1: $ Reference distribution (e.g. set of embeddings)
    \State $D_2$: Target distribution (e.g. set of embeddings)
     \State \textbf{Variable:}

	\State $\beta:$ Number of samples to be compared
	\State $M:$ Total number of samples

    \State $K: $ Number of bootstraps
    \State $D_{1,j}:$ $j$th sample of reference distribution
     \State $D_{2,j}:$ $j$th sample of target distribution
    \State $R:$ List of MMD statistics
    \State $R_b:$ List of bootstraped MMD statistics 
    \State $X:$ List of Median MMD statistics
	


  \State $t:$ Index
	\For{$t=\beta:M$}
	
		\State  $Q_1=\{D_{1,1},D_{1,2},...,D_{1,t}\}$ \Comment{batch of reference distribution}
			\State  $Q_2=\{D_{2,1},D_{2,2},...,D_{2,t}\}$ \Comment{batch of target distribution}
		\State $r=MMD(Q_1,Q_2)$ \Comment{compute Maximum Mean Discrepancy}
			\State $T=\{Q_1,Q_2\}$  \Comment{combine $Q_1$ and $Q_2$ (under  $H_0$ of no drift ) }
   	\State $R=\{R;r\}$ \Comment{append MMD statistic}
	\State \For{ $i=1:K$}
	\State $T^{'}=Bootstrap(T)$  \Comment{get a bootstrap samples from $T$}  
	\State $Q_1^{'}=T^{'}_{1:\beta/2}$  \Comment{get 1st half of $T^{'}$}
	\State $Q_2^{'}=T^{'}_{\beta/2+1:\beta}$  \Comment{get 2nd half of $T^{'}$}
	
	\State  $r^{'}=MMD (Q_1^{'},Q_2^{'})$ 
	\State $R_b=\{R_b;r^{'}\}$ \Comment{append MMD statistic}
	\EndFor
	
     \State $d_{med}=\textit{Median}(R_b)$ \Comment{compute median of estimated MMDs}
	
	
	\State $X=\{X;d_{med}\}$ \Comment{append $d_{med}$ to $X$}
	
		\State $t+=1$  \Comment{increase index}
		
        \State $Q_1=\{Q_1;D_{1,t}\}$ \Comment{add next sample to $Q_1$}
        
     	\State $Q_2=\{Q_2;D_{2,t}\}$ \Comment{add next sample to $Q_2$}
	
	   \State $Q_1.\textit{popleft()}$ \Comment{remove first sample from $Q_1$}
        
     	\State $Q_2.\textit{popleft()}$ \Comment{remove first sample from $Q_2$}

	\EndFor
	  \State $z$=$Argmax(R)$ \Comment{highest drift Index}
   \State $U = \{D_{l,z-\beta},D_{l,z-\beta+1},...,D_{l,z}\}\; \; \; l\in \{1,2\}$  
	\State Return $d=\textit{mean} (X)$ \Comment{median MMD}
    \State  Return $U$  \Comment{samples causing highest drift}
    
	\end{algorithmic}

\end{algorithm}\label{al:al1}

\section{Experiment and Results}

This section showcases the effectiveness of our proposed drift detection methodology in three areas: (1) detecting model performance regression, (2) implementing mitigation strategies for model performance regression, and (3) text encoder effect ablation study. The overall approach that we present here can be applied to any domain or dataset.

\subsection{Model Performance Regression Detection}

To demonstrate a significant negative correlation between model performance metrics and estimated drift from our proposed drift detection method, we conducted an experiment based on a binary classification model, where the model is designed to detect if a text sentence is related to shopping in the categories such as searching, buying, and checking price for items. For this binary classification model, we use BERT \citep{devlin2018bert} encoder to obtain the embedding of a sentence and then the embedding goes through three feed forward dense layers for binary classification task. The model was trained on a dataset comprising approximately 800K annotated samples and evaluated on a separate test set of around 150K data points.

\begin{figure}[h]
  \centering
  \includegraphics[scale=0.40]{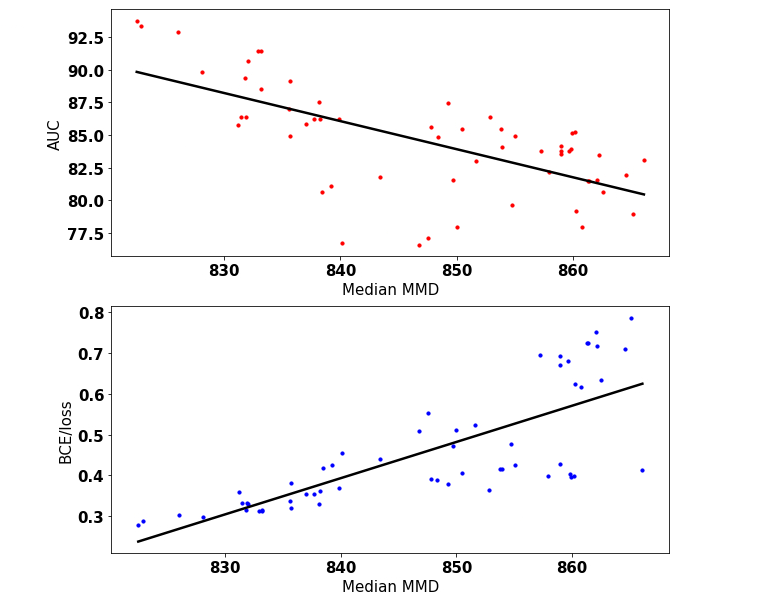}
  \caption{Plot of estimated drift (MMD) vs the model performance metrics for monthly buckets, which indicates that as the amount of drift increases, model performance drops.}
  \label{fig:1}
  
\end{figure}

After training and tuning the model on validation data, we evaluated its performance on test data. Then, we conducted an experiment to examine the relationship between its performance metrics and drift estimated by our proposed drift detection method. For this experiment, we collected data for about 3 years and divided it into monthly buckets. We then calculated the amount of drift between the model's development data and the data from each monthly bucket, and also computed the model's area under the ROC curve (AUC) and binary cross entropy (BCE) for each bucket as performance metrics. In our drift detection method, we used the BERT encoder for getting the embedding of text data \citep{devlin2018bert}. The results of this experiment are presented in Figure \ref{fig:1}, which clearly shows that as the amount of drift increases, the model's performance significantly decreases.

\begin{table}[H]
    \centering
    \begin{tabular}{|c|c|}
    \hline
    Metric     & correlation (\%) \\
    \hline
      MMD vs BCE   & 76.9\\
      \hline
      MMD vs AUC & -65.2\\
      \hline
    \end{tabular}
    \caption{Correlation coefficient between estimated drift (MMD) and model performance metrics.}
    \label{tab:cor}
\end{table}

\begin{figure}[h]
  \centering
 
    \includegraphics[scale=0.48]{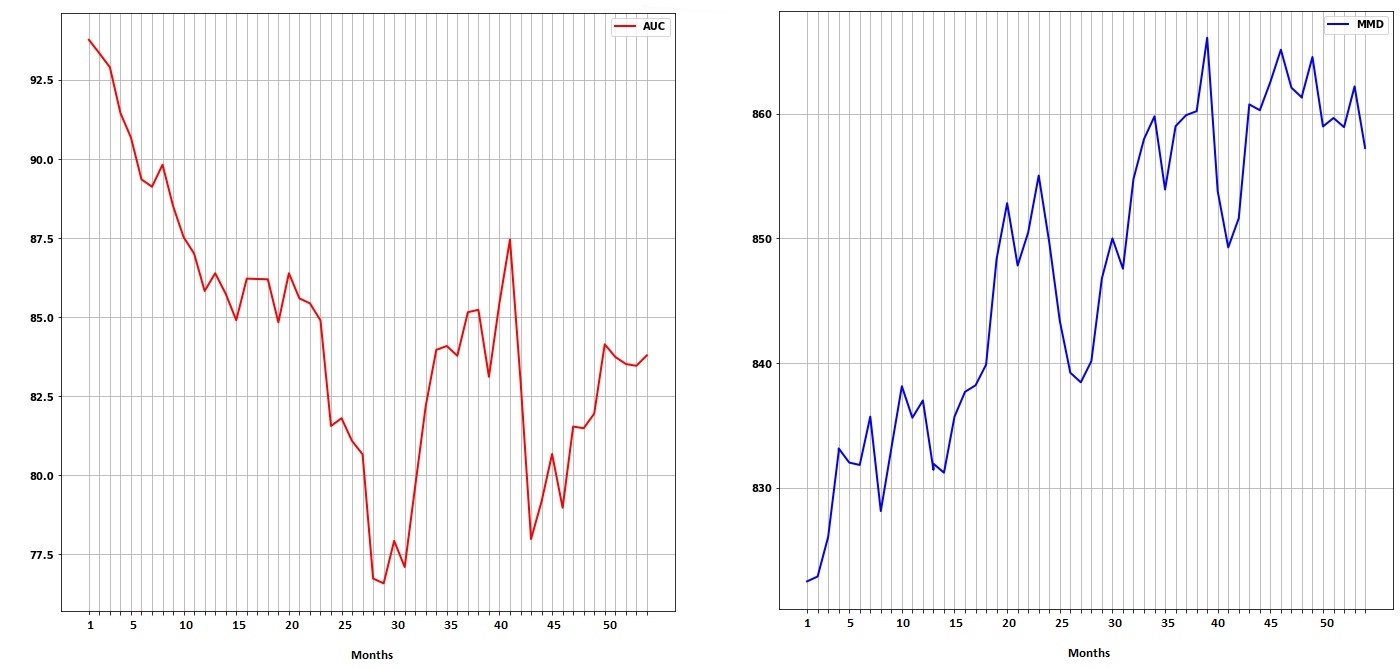}
 
  \caption{Left and right plots shows the model performance metric (AUC) and the estimated amount of drift (MMD), respectively. The increase in MMD over time is attributed to the emergence of new data patterns in the production data, causing an increase in the estimated drift and a corresponding drop in the model’s performance}
  \label{fig:twosidebyside}
\end{figure}

The results in Table \ref{tab:cor} demonstrate a significant correlation between estimated drift (MMD) and model performance metrics. Consequently, our unsupervised drift detection approach can effectively monitor model performance and accurately predict model regression. This capability is particularly valuable in production environments where annotated data may not be available, making our drift detection method the only practical option for model performance monitoring. Figure \ref{fig:twosidebyside} depicts the estimated drift and model performance over time, revealing a notable increase in drift as time progresses. This increase is attributed to the emergence of new data patterns in the production data, causing an increase in the estimated drift and a corresponding drop in the model's performance. These findings underscore the need for ongoing monitoring and recalibration of the model to ensure its continued effectiveness.

\subsection{Model Performance Regression Mitigation}

To assess the effectiveness of our proposed drift detection method in reducing the decline in model performance from drift when used in production, we outline a comprehensive mitigation process. Firstly, we run our drift detection method by setting the reference and target distributions to be the model's training and inference data during production, respectively. Then, our proposed method identifies the samples that exhibit the highest degree of drift or deviation from the production data. Finally, we incorporate the dataset generated from the previous step into the model's training data and retrain the model, enabling it to improve its performance during production.

We implemented the above-mentioned approach on a multi-task model used for domain classification, intent classification, and name entity recognition. The model architecture comprises a blend of Bi-LSTM and transformer models \citep{devlin2018bert}.

To improve this model performance, we employed our proposed drift detection method to compare the model's training data with one month of production data. This enabled us to identify samples with the highest degree of drift from production data. However, in order to use these data during model training, they need to be annotated. Therefore, we utilized a semi-supervised approach, where we used a large transformer model that leverages a variety of offline signals such as previous conversations that are not available during online inference, to generate pseudo-labels for these data. With the help of these pseudo-labels, we added new samples with highest amount of drift to the model's training data and retrained the model.

We conducted a thorough evaluation of the new model trained on samples generated from our drift detection method. Specifically, we tested the model's performance on a held-out false accept (FA) dataset, consisting of only false accept data, to evaluate its ability to reject false accept samples. Additionally, we compared the performance of our proposed approach with several common mitigation methods. These methods include the following:

\textbf{Baseline:} A baseline model trained using the model's original training data without any mitigation.

\textbf{Bias reduction:} This method involves clustering the training and production data to identify patterns that are absent within the training data. We then added new samples from absent patterns to the training data using this approach.

\textbf{Upsampling:} This method first identifies the underrepresented samples within the training data by clustering and then employs upsampling to increase their frequency in the training data. Then, we added underrepresented samples to the training data using this approach.

We re-trained the model using all these methods, and Table \ref{tab:test} provides a detailed comparison of the various methods used in this experiment. The results indicate that the samples of data added by our proposed drift detection method significantly improve model performance and outperform other methods to varying degrees, without an increase in false reject rate.

\begin{table}[H]
    \centering
    \begin{tabular}{|c|c|}
    \hline
    Method     & FAR (\%) \\
    \hline
      Baseline   & 73.15\\
      \hline
      Bias reduction & 61.85\\
      \hline
         Upsampling & 71.48\\
      \hline
      
         Proposed drift detection  & \textbf{59.68}\\
      \hline
    \end{tabular}
    \caption{Comparison of different mitigation methods on false accept rate (FAR) test set. Our proposed drift detection method shows superior performance compared to the other methods. Lower FAR values indicate better performance.}
    \label{tab:test}
\end{table}

\subsection{Encoder Effect Ablation Study}

In this section, we outline an experiment designed to assess the effect of different encoders on the performance of our proposed method. We utilize different encoders to extract text embeddings and focus on simulating data drift within a binary classification scenario. The reference dataset initially maintains a balanced distribution of positive and negative class instances, each accounting for 50\% of the dataset. Subsequently, we change the positive class percentage within the target dataset to induce data drift. Following datasets are used in our experiment:

\textbf{AG news:} AG news comprises over one million news articles, compiled from a diverse range of 2000+ news sources over a span of more than a year \citep{Zhang2015CharacterlevelCN}. We use world and sports classes in our experiment.

\textbf{Yelp review:} yelp review dataset consists of reviews extracted from the Yelp \citep{zhang2016characterlevel}. We use reviews rated 5 as the positive class and reviews rated 1 as the negative class in our experiment.

We sample 5000 observations from each dataset where we increase/decrease the percentage of positive class in the target dataset while comparing it to the reference dataset with equal ratio of positive and negative classes. We set the batch size, $\beta$, and bootstrap steps of our method to be 64, 32, and 50, respectively. We use the following encoders in our experiment:

\textbf{bert-base-uncased:} BERT base \citep{devlin2018bert} is a transformer encoder model with 12 layers with 768 hidden size that is pretrained on a large corpus of English data.

\textbf{bert-large-uncased:} this is a BERT model with 24 layers with 1024 hidden size.

\textbf{all-MiniLM-L12-v2:} This model, known as a sentence-transformers model, effectively encodes sentences and paragraphs into a 384-dimensional dense vector space, offering utility in applications such as clustering or semantic search \citep{sbertnet}. This model has 12 layers with 384 hidden size.

\textbf{all-distilroberta-v1:} This model is also a sentence-transformer with 12 layers and 768 hidden size.

We ran our drift detection method using all four different encoders on both datasets and the following Figure \ref{fig:twosidebyside_yelp} shows the results. The results indicate that there is a small amount of drift when the percentage of positive class is close to 50\% which is the exact distribution of the reference dataset. However, the amount of drift increases as the percentage of positive class gets more different than the reference distribution. All encoders performed comparably but the sentence BERT and larger transformer models tend to detect drift faster due to producing higher quality text representations.

\begin{figure}[h]
  \centering
 
    \includegraphics[scale=0.45]{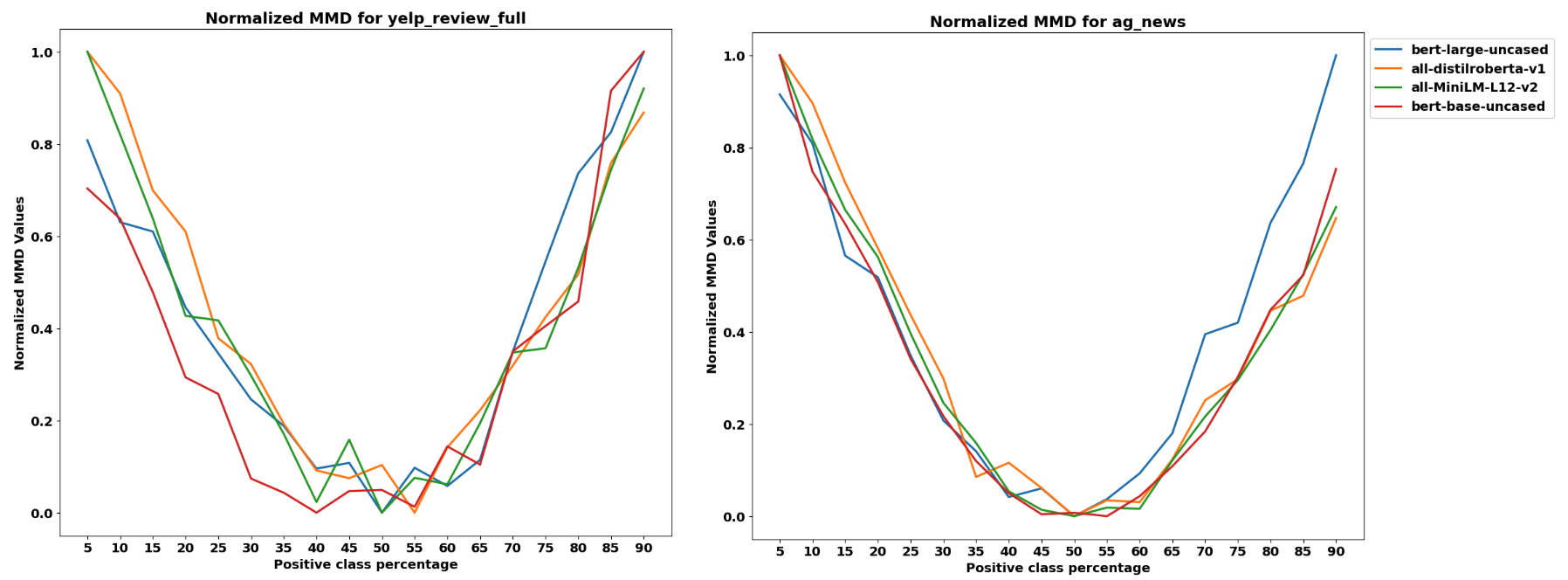}
 
  \caption{Left and right plots shows MMD values as we change the percentage of positive class in the target data resulting in data drift.}
  \label{fig:twosidebyside_yelp}
\end{figure}

\section{Discussion and Conclusion }

Our proposed unsupervised drift detection approach for unstructured data, specifically text data, presents a novel and effective method for identifying potential drift in machine learning models. By utilizing maximum mean discrepancy (MMD) kernel-based statistical test and bootstrap, we can accurately identify the subset of data likely to be the root cause of the drift and suggest its adding back to the training data as a mitigation strategy. Our approach is particularly useful in identifying model performance regression and provide a fast mitigation solution for recovery.

In this paper, we demonstrated the efficacy of our proposed method for two use cases, but the approach can be applied to any other domain or dataset. Overall, our proposed method provides a robust framework for detecting and mitigating performance regression in machine learning models caused by drift.

\bibliographystyle{unsrtnat}
\bibliography{references}  






\end{document}